\documentclass[letterpaper, 10 pt, conference]{ieeeconf}  

\IEEEoverridecommandlockouts                              

\usepackage{amsmath}
\usepackage{amssymb}
  
\usepackage{amsthm}
\usepackage{graphicx}
\usepackage[Symbol]{upgreek}
\usepackage{epstopdf}
\usepackage{subcaption}
\usepackage{enumerate}
\usepackage{paralist}

\usepackage{float}

\usepackage{latexsym}
\usepackage{multicol}
\usepackage{multirow}
\usepackage{lipsum}

\usepackage{cite}

\usepackage{makecell}
\usepackage{breqn}

\usepackage{gensymb}

\usepackage{verbatim}

\usepackage{color}
\usepackage{xcolor}
\usepackage{bm}
\usepackage[font=small,labelfont=bf]{caption}

\usepackage{courier}
\usepackage{tikz}
\usetikzlibrary{calc}

\newtheorem{prop}{Proposition}

\def\particulartemplate#1{
	\begin{tikzpicture}[overlay, remember picture]
	\draw let \p1 = (current page.west), \p2 = (current page.east) in
	node[minimum width=\x2-\x1, minimum height=0.1cm, rectangle, fill=yellow!35!white, anchor=north west, align=center, text width=\x2-\x1] at ($(current page.north west) + (0,-0.3)$) {\large \textbf{\texttt{#1}} };
	\end{tikzpicture}
}

\newcommand{\vct}{\mathbf}
\newcommand{\vect}[1]{\boldsymbol{#1}}

\title{\LARGE \bf
A controller for reaching and unveiling a partially occluded object of interest with an eye-in-hand robot  
}

\author{Dimitrios Papageorgiou, Leonidas Koutras and Zoe Doulgeri
\thanks{The research leading to these results has received funding from the European Community’s Framework Programme Horizon 2020 under grant agreement No 871704, project BACCHUS.
}
\thanks{Authors are with the Automation \& Robotics Lab, Dept.
of Electrical \& Computer Engineering, Aristotle University of
Thessaloniki,
Greece {\tt\small \{dimpapag, kleonidas, doulgeri\}@ece.auth.gr}}

}

\begin{document}

\maketitle
\particulartemplate{
	This paper is a pre-print version under review. \\
}
\thispagestyle{empty}
\pagestyle{empty}


\begin{abstract} \label{sec:abstract}
In this work, a control scheme for   approaching and unveiling  a partially occluded object of interest is proposed. 
The control scheme is based only on the classified point cloud obtained by the in-hand camera attached to the robot's end effector. It is shown that the proposed controller reaches in the vicinity of the object  progressively  unveiling the neighborhood of each visible point of the object of interest.  It can therefore potentially achieve the complete unveiling of the object. The proposed control scheme is evaluated through  simulations  and experiments with a UR5e robot with an in-hand RealSense camera on a mock-up vine setup for unveiling the stem of a grape.   
\end{abstract}


\section{Introduction} \label{sec:Introduction}

As  opposed  to  the  structured  industrial environment,  farming  fields  are  unstructured  and  the objects  of  interest (OOI),  e.g.  fruits,  stalks, or  stems,  are in  many  cases either  partially  or  completely occluded, when the robot attempts to acquire information related to them. Furthermore, in most cases,  the  fruit  cannot  be  concretely  modelled,  as  it  involves  structural  uncertainties and its location  is  not  known a  priori\cite{Zapotezny2019}.  Therefore,  the  challenge  of visually  unveiling the  OOI, by tackling the problem of avoiding possible occlusions, e.g. leaves or branches, is widely reported and is considered as a non-trivial issue in agricultural robotics\cite{Zapotezny2019,Lehnert2019,Gongal2015}.

In this work we consider the problem of approaching and unveiling a partially occluded object of interest to enable a subsequent task of grasping or cutting. We consider a  robot with a camera in hand and design a control scheme that requires a perception system  that can on-line classify the obtained points of the cloud to those belonging to the object of interest and the rest of the objects that potentially occlude it. The method does not require models of the object of interest and the surrounding obstacles and it can thus be used in agricultural applications. There is further no need  to resort to task space exploration motions  space for unveiling it.

\section{Related literature}

The methods for visually unveiling the object of interest, by addressing the problem of occlusions, considering  a  robot  with  an  in-hand  camera,  can  be  divided  in  three  main  categories:  the RGB image based optimization approaches, the 3D scene geometry based approaches and the machine learning based approaches.

\subsection{RGB image-based optimization approaches}

According   to   the optimization-based approaches, which   exploit  only   RGB   information, an optimization function is defined, and the motion of the robot aims at maximizing (or minimizing) this function. In the context of visual unveiling of the OOI, one commonly selected optimization function is the visible area of the object of interest, i.e. the number of pixels of the OOI within the image captured  by the  camera. More  specifically, in \cite{Lehnert2019}, an  agricultural harvesting scenario  is considered  and the  optimization  function  involves  the  pixels  of  the  visible  OOI,  as  well  as a manipulability  measure. However,  for  finding  the  gradient  of  the  optimization  function, which defines the direction towards the function’s maximization, 9 cameras placed in a 3x3 grid-fixture  were utilized and attached to the robot’s end-effector. On the other hand, methods relying on a single camera for maximizing an optimization function involve either a predefined scanning path to obtain multiple views\cite{Roy2004}, which may be time and energy consuming, or probing strategies for exploration\cite{Farrokhsiar2013}, which yield suboptimal solutions.

\subsection{3D scene geometry-based approaches}

Methods  that  utilize  information  about  the  3D  scene  geometry  are  the  most  popular  solutions amongst the robotics literature. According to such approaches, the control signal which actuates the  in-hand  camera  is  based  on  the information  about  the 3D geometry  of  the  environment, acquired via appropriate sensors, e.g. RGB-D cameras or LiDARs.

Most of the methods of this category assume a known and modelled environment and/or involve a known target (i.e. an object with known and predefined geometry/structure) \cite{Penin2018,Cowan2002,LaValle1997,Baumann2008,Kazemi2009,Nicolis2018,Cuiral2020}. More specifically, in \cite{Penin2018}, an autonomous drone equipped with a camera is considered, with the obstacles being modelled by spheres and the target (i.e. the OOI) being a point-feature having a known position. In \cite{Cowan2002}, the authors propose a control scheme for self-occlusion avoidance during visual servoing, considering a complete knowledge of the object’s model. In \cite{LaValle1997}, a trajectory planning algorithm for reaching a specific target is presented, which assumes that the complete occlusion-free  space is known  and/or is computable a  priori. Similarly, in \cite{Baumann2008}, the  complete geometry of the obstacles (clutter) within the environment is assumed to be known, e.g. by the CAD  in  an  industrial  task, based  on  which the  occlusion-free  space  is analytically calculated, considering a sphere-like visual target with a known position. Following this line of thought, also the approaches of \cite{Kazemi2009} and \cite{Nicolis2018} assume that the 3D model of the target object and the obstacles in the workspace are known a priori. Lastly, in \cite{Cuiral2020}, although the obstacles are not modelled, the normal vectors at each point of the surface of the OOI are assumed to be known or computable, which implies that the OOI surface is either pre-modelled or can be modelled on-line. 

In \cite{khelloufi2020},  a  method  which  does  not involve  the  modelling  of  the  environment  or  the  target  is proposed. However, it is applicable only in the case of a 2D navigation of a platform, as a number of virtual 1-dof curves are defined around the robotic platform, having a geometry which cannot be directly extended to the 3D space. 

\subsection{Machine learning based approaches}

A machine learning approach is proposed in \cite{Zapotezny2019}, built on top of \cite{Lehnert2019}. In particular, as opposed to \cite{Lehnert2019}, a convolutional  neural  network  (CNN) is  utilized in\cite{Zapotezny2019}, to estimate the  gradient of the optimization function, making the method also applicable by utilizing only a single camera setup, instead  of 9  cameras required in\cite{Lehnert2019}. However, the  training  of  the  CNN  requires  a  large  and representative training dataset.   Although   this   dataset   can   be gathered via   a   simulated environment, transferring the trained controller to the field is still a challenging task. In \cite{Morrison2019}, an eye-in-hand grasping task is considered and a CNN is utilized for calculating a grasping feasibility map. However, to find the “next best view”, the method requires the exploration of the task space (e.g.  to  pass  through  various  viewpoints), and the calculation  of the grasping map at  each view points of the task space, making this method time and energy consuming. 

Finally, a work which cannot be easily classified in one of the aforementioned categories is \cite{Ourak2019}, in  which  a  wavelet  transformation  is  utilized  in  order  to perform  a  visual  servoing  task  by avoiding occlusions. However, this method requires the knowledge of the exact pattern or model of the OOI, to assess the similarity via the wavelet transformation. Unfortunately, this information is usually not available in robotically assistive farming applications.

\section{  Problem description} \label{sec:prob_description}

Consider the availability of an N-dof robotic manipulator with an RGB-D camera attached to its end-effector. Let $\vct{q}\in\mathbb{R}^N$ be the
vector of the joint variables of the manipulator, with $N$ being the number of joints. In the case of a mobile manipulator the degrees of freedom include those of the mobile platform (3 dof). In the rest of the paper,  most variables are  expressed in the camera frame. Expressions in other frames (e.g the world frame) will be denoted by a left superscript.

Let ${}^0\vct{T}_c\in SE(3)$ be the homogeneous transformation expressing the generalized pose of the camera frame \{C\} in the
 world inertial frame \{0\}, involving its position ${}^0\vct{p}_c(\vct{q})\in\mathbb{R}^3$  and its orientation ${}^0\vct{R}_c(\vct{q})\triangleq [\vct{x}_c \; \vct{y}_c \; \vct{z}_c ] \in SO(3)$. Let us further denote by $\vct{v}_c$ the generalized body velocity of the camera's frame consisted of the translational velocity ${}^0 \vct{R}_c^\intercal {}^0 \dot {\vct{p}}_c$ and the rotational velocity $\vct{S}(\vect{\omega}_c) ={}^0 \vct{R}_c^\intercal {}^0 \dot{\vct{R}}_c $, with $\vct{S}(.)$ denoting the skew-symmentric matrix.
 We assume that the
 system is aware of the center of the region of interest (ROI) that 
includes at least one object of interest (OOI), that should not necessarily be visible by the robot when it is away from the ROI. For example, a grape cluster in the center of ROI has an associated stem that may not be visible at start. 

We model a ROI  by a
sphere; let $\vct{p}_r\in\mathbb{R}^3$ be the center of the ROI
and $r\in\mathbb{R}_{>0}$
be its radius.
We assume that given the point-cloud perceived by
the RGB-D camera, the perception algorithm can on-line  classify the points of the cloud in two categories: a) those belonging to the OOI and b) the rest of the point-cloud which may belong to surfaces that occlude the OOI. For instance, in the
grape harvesting case, the OOI is the stalk of the grape cluster, while the rest of the point cloud
may be points of the leaves, branches, supporting wires or even the rest of the fruit. Let $\vct{p}_i\in\mathbb{R}^3,\;i=1,...,n$ be the currently visible points of the OOI and $\vct{p}_{o,j}\in\mathbb{R}^3,\;j=1,...,m$ be the rest of
the points which belong to surfaces that may occlude the observation of the OOI.

We assume a robot that provides a position or velocity control interface which is true for most of the commercially available collaborative robots.  
In the kinematic level, the controlled system can  be expressed  in the joint space as follows:
\begin{equation}\label{eq:controlled_system}
    \dot{\vct{q}}=\vct{J}^\dagger(\vct q) \vct{u}_c,
\end{equation}
where $\vct{J}^\dagger\in\mathbb{R}^{N\times6}$ is a pseudo inverse of the end-effector Jacobian matrix $\vct{J}(\vct{q}) \in \mathbb{R}^{6\times N}$ and $\vct{u}_c\in\mathbb{R}^6$ is the control input that takes the form of the body reference velocity of the camera's frame and which  should be designed in order to  achieve the
following objectives:

\begin{enumerate}
    \item The end-effector has to smoothly reach the region of interest (ROI), which can be
mathematically expressed as: $\|\vct{p}_r(t)\|\rightarrow\overline{r}\leq r$, for $t\rightarrow\infty$, for some scalar positive $\overline{r}$. \label{objective:region_reaching}
    \item To maintain the center of ROI at the center of the field of view, as much as possible, which
can be mathematically expressed as: $\theta(\vct{p}_r)\rightarrow 0$, for $t\rightarrow\infty$, where $\theta(\vct{p}_r)\triangleq\text{cos}^{-1}\left(\frac{\vct{z}^\intercal\vct{p}_r}{\|\vct{p}_r\|}\right)$ is the angle between  $\vct{p}_r$ and the   camera's view direction  which we assume without loss of generality that coincides with the z-axis of the camera's frame   $\vct{z}=[0 \; 0 \;1]^\intercal$. Notice that $\theta$ cannot be more than $\frac{\pi}{2}$, since the center of ROI $\vct{p}_r$ is within the
field of view of the camera, i.e. a pyramid.  \label{objective:centering}
    \item To maximize the visible area of the object of interest, i.e. to maximize the perceived number
of points of the point-cloud of OOI. \label{objective:occlusion_avoidance}
\end{enumerate}

Objective \ref{objective:region_reaching} aims at moving the end-effector to any of the points of the
ROI while objective \ref{objective:centering} aims at aligning the $z$-axis of the camera with the vector pointing to the center of ROI.
Clearly these two objectives can be achieved by reaching any point in ROI with 
specified two orientation degrees of freedom
(DOF) of the camera frame. Given the task  redundancy
the third
objective is set to select the point in ROI that visually unveils the OOI as much as possible.

\section{  Proposed control scheme} \label{sec:control_scheme}

The controller superimposes two
control signals $\vct{v}_{c1}$, $\vct{v}_{c2}$ as follows:
\begin{equation}\label{eq:superimposition}
    \vct{u}_c=\vct{v}_{c1} + \vct{v}_{c2},
\end{equation}
in the form of end-effector body velocities  which
are the following:
\begin{itemize}
    \item $\vct{v}_{c1}\in\mathbb{R}^6$: ROI reaching \& centering control signal, which is designed to fulfill the first two
control objectives (Objective \ref{objective:region_reaching} and Objective \ref{objective:centering}).
    \item $\vct{v}_{c2}\in\mathbb{R}^6$: OOI active perception control signal, which is designed to fulfill the last
objective (Objective \ref{objective:occlusion_avoidance}), which is related to the maximization of the visible area of the OOI.
\end{itemize}
Each signal, acting
alone, aims at achieving the respective objective.

The proposed control scheme is designed so that $\vct{v}_{c2} = \vct{0}$ when no points of the OOI are visible, i.e. $n=0$. Then  the ROI reaching \& centering controller is acting alone. Such a case may occur during the start of the
motion as we mentioned before when the camera is far from the ROI and the OOI starts being visible only when it
approaches or enters the ROI.

\subsection{ROI reaching \& centering control signal} \label{subsec:rr_control_signal}

Notice that control objectives \ref{objective:region_reaching} and \ref{objective:centering} can be achieved by appropriately translating and rotating the end-effector camera. 
Mathematically, the first objective is
fulfilled when the camera position $^0\vct{p}_c$ converges to the manifold $^0\Omega$:
\begin{equation} \label{eq:omega}
    ^0\Omega\triangleq\{{}^0\vct{p}_c\in\mathbb{R}^3:f({}^0\vct{p}_c-{}^0\vct{p}_r)\leq 0\},
\end{equation}
with $f(\vct{x})\triangleq\vct{x}^\intercal\vct{x}-r^2$ for any $\vct x \in \mathbb{R}^3$. Since ${}^0\vct{p}_r={}^0\vct{R}_c\vct{p}_r+{}^0\vct{p}_c$ the manifold \eqref{eq:omega} condition can be expressed in the camera frame {C} as follows:
\begin{equation} \label{eq:obj_reaching_ineq}
    f(\vct{p}_r)\leq0,
\end{equation}

Objective \ref{objective:centering} regards the orientation of the camera. It is achieved when the camera's $z$-axis points to the center of the ROI.

To achieve these objectives, we propose the following control signal, which draws its inspiration
from the region reaching approach proposed in \cite{Cheah2005} for the translational part:
\begin{equation} \label{eq:control_signal_rr}
    \vct{v}_{c1}\triangleq 
    \begin{bmatrix}
       k_p\text{max}(0,\; f(\vct{p}_r))\vct{p}_r \\
       k_o \theta \vct{k}
    \end{bmatrix},
\end{equation}
where $k_p, k_o\in\mathbb{R}_{>0}$ are constant positive gains for translation and orientation respectively and $\theta\in[0,\frac{\pi}{2}]$ and $\vct k$ are the angle and
axis of the minimum rotation between $\vct{p}_r$ and $\vct{z} \triangleq [0 \; 0 \; 1]^\intercal$, which can be calculated by the
following expressions:
\begin{equation} \label{eq:ktheta}
    \theta = \text{cos}^{-1}\left(\frac{\vct{z}^\intercal\vct{p}_r}{\|\vct{p}_r\|}\right), \; 
    \vct{k} = \frac{\vct{S}(\vct{z})\vct{p}_r}{\|\vct{S}(\vct{z})\vct{p}_r\|}, 
\end{equation}
where $\vct{S}(\vct{z})$ is the skew symmetric matrix of the unit vector $\vct{z}$.

Notice that $\vct{v}_{c1}$ is continuous and its value is zero if and only if
$f\leq 0$ and $\theta = 0$. 
\begin{prop}
    Setting   $\vct u_c=\vct{v}_{c1}$ in \eqref{eq:controlled_system} achieves asymptotic convergence to a point that fulfills Objectives \ref{objective:region_reaching} and \ref{objective:centering}. 
\end{prop}
The proof can be found in the Appendix. 


\subsection{ Visually unveiling OOI active perception control} \label{subsec:occlusion_control_signal}

\begin{figure}[!htbp]
	\centering
	\includegraphics[scale=0.35]{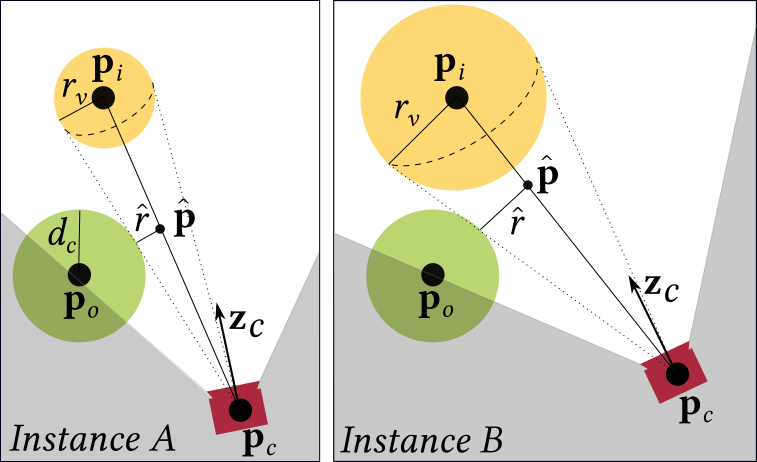}
	\caption{Basic concept involving a single point of the OOI and a single point as an obstacle.}
	\label{fig:motivation}
\end{figure}

Assume the simple case depicted in Figure \ref{fig:motivation}. Let $\vct{p}_i$ be a visible point of the object of interest (OOI). Let  $\vct{p}_o$ be the center of a spherical obstacle that can potentially occlude the OOI for a camera pose. The ray from the camera $\vct p_c$ to $\vct{p}_i$ shown Figure \ref{fig:motivation} is in the camera's field of view since $\vct{p}_i$ is visible. The vector normal to this ray from  $\vct{p}_o$ intersects the ray at point  $\hat{\vct{p}}$.
Let $\hat{r}\in\mathbb{R}_{\geq0}$ be the distance along the normal from the obstacle surface to $\hat{\vct{p}}$. It is clear that the maximum visible spherical region centered at $\vct{p}_i$ has a radius $r_v$ which is always greater or equal to $\hat{r}$.
Comparing the instances A and B of Figure \ref{fig:motivation}, notice that if the distance of the camera from $\vct p_i$ remains constant or less, the greater the value of $\hat{r}$, the greater the volume of the neighborhood of $\vct{p}_i$ that is unveiled.

Based on this observation, we design a control signal for the camera's body velocity $\vct{v}_{c2}$ which can be seen as a rotation of the camera around $\vct{p}_i$ in order  to  guarantee that $\hat{r}>d_c$, and that its derivative is positive, i.e., that $\hat{r}$ will increase so that  the unveiled region around  $\vct{p}_i$ will increase.

 Consider the general case of $\vct{p}_i\in\mathbb{R}^3, i=1,...,n$ currently visible points and $\vct{p}_{o,j}\in\mathbb{R}^3, j=1,...,m$ obstacles. Given the aforementioned idea and drawing our inspiration from our previous work \cite{Kastritsi2019}, we propose the utilization of a barrier artificial potential field around each obstacle, which induces a virtual repulsive velocity $\vct{u}_{i,j}$, acting at $\hat{\vct{p}}_{i,j}$, which is the nearest point of the $i$-th ray from the $j$-th obstacle and is calculated by:
 \begin{equation}\label{eq:nearest}
    \hat{\vct{p}}_{i,j}\triangleq \begin{cases}
        \vct{0}, \; &\text{if} \; \frac{\vct{p}_i^\intercal}{\|\vct{p}_i\|} \vct{p}_{o,j} \leq 0 \\
        \frac{\vct{p}_i\vct{p}_i^\intercal}{\vct{p}_i^\intercal\vct{p}_i} \vct{p}_{o,j}, \; &\text{if} \; \frac{\vct{p}_i^\intercal}{\|\vct{p}_i\|} \vct{p}_{o,j} \in(0, \|\vct{p}_i\|) \\
        \vct{p}_i, \; &\text{if} \; \frac{\vct{p}_i^\intercal}{\|\vct{p}_i\|} \vct{p}_{o,j} \geq \|\vct{p}_i\| \\
    \end{cases}.
\end{equation}

\begin{figure}[!htbp]
	\centering
	\includegraphics[scale=0.6]{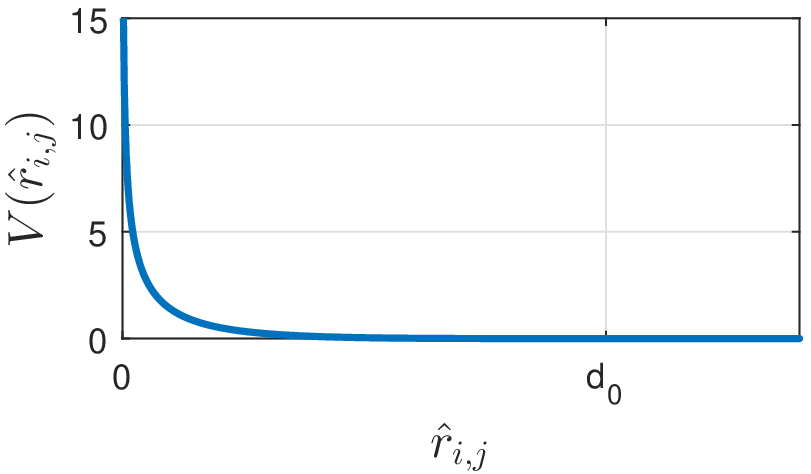}
	\caption{Artificial potential function for $\vct{v}_{v2}$.}
	\label{fig:potential_vc2}
\end{figure}

The proposed barrier artificial potential function,  depicted in Figure \ref{fig:potential_vc2}, is designed to induce a repulsive velocity only within a predefined distance $d_0\in\mathbb{R}_{>0}$ from the obstacle surface; it is defined as follows:
\begin{equation}\label{eq:art_potential}
    V(\hat{r}_{i,j})\triangleq 
    \begin{cases}
    \frac{1}{2}\ln\left(\frac{d_0^2}{d_0^2-(d_0-\hat{r}_{i,j})^2}\right)^2, \; &\text{if} \; \hat{r}_{i,j}< d_0 \\
    0, \; & \text{otherwise} 
    \end{cases}
\end{equation}
where
\[
\hat{r}_{i,j}\triangleq \|\hat{\vct{p}}_{i,j}-\vct{p}_{o,j}\|-d_c
\]
with $d_c\in\mathbb{R}_{>0}$ the obstacle's spherical radius that is selected for minimizing the coverage of the empty space between neighboring points of the obstacle cloud.   
Notice that the value of the artificial potential  tends to infinity when the surface of the obstacle is reached from outside in order to guarantee obstacle/occlusion avoidance.

The virtual commanded repulsive velocity is then given by
\begin{equation}\label{eq:u_ij}
\begin{split}
    &\vct{u}_{i,j} \triangleq - \frac{\partial V(\hat{\vct{p}}_{i,j})}{\partial\hat{\vct{p}}_{i,j}} = \\
    &\begin{cases}
    \frac{2}{d_0^2-(d_0-\hat{r}_{i,j})^2}\ln\left(\frac{d_0^2}{d_0^2-(d_0-\hat{r}_{i,j})^2}\right)\vct{e}_{i,j}, \;  &\text{if} \; \hat{r}_{i,j}< d_0 \\
    \vct{0}_3, \; &\text{otherwise} 
    \end{cases}
    \end{split}
\end{equation}
where $\vct{e}_{i,j}\in\mathbb{R}^3$ is given by:
\begin{equation}\label{eq:e_ij}
    \vct{e}_{i,j}\triangleq (d_o-\hat{r}_{i,j})\frac{\hat{\vct{p}}_{i,j} - \vct{p}_{o,j}}{\|\hat{\vct{p}}_{i,j} - \vct{p}_{o,j}\|}.
\end{equation}
which is a vector at the direction of the normal to the ray with magnitude proportional to the distance along the normal from the obstacle's surface.

The virtual repulsive velocity \eqref{eq:u_ij} possesses the following properties:
\begin{itemize}
    \item $\|\vct{u}_{i,j}\|\neq 0$, if and only if $0<\hat{r}_{i,j}<d_0$, and $\vct{u}_{i,j}\neq\vct{0}$ if and only if $\hat{r}_{i,j}\geq d_0$, which means that the signal will not be affected by rays that are not within the range of influence of the $j$-th obstacle point, defined by $d_0$.
    \item $\|\vct{u}_{i,j}\|\rightarrow \infty$, when $\hat{r}_{i,j}\rightarrow 0$, i.e. when the $i$-th ray approaches the  surface of the $j$-th obstacle. Notice that $\|\hat{\vct{p}}_{i,j}-\vct{p}_{o,j}\|$ cannot be less than $d_c$, as it reflects the accuracy of the RGB-D camera, by definition.
    \item $\vct{u}_{i,j}$ is continuous with respect to $\hat{\vct{p}}_{i,j}$.
    \item However, $\vct{u}_{i,j}$ is, in general, not continuous in time, as $\vct{p}_i, \vct{p}_{o,j}$ and even $n,m$ depend on the point-cloud perceived by the RGB-D camera during its motion; as more points of the OOI, previously occluded, enter into the field of view, it is possible that some of them induce a non-zero control signal. This discontinuity can be remedied by a first order low-pass filter.
\end{itemize}

To synthesize the total proposed control signal $\vct{v}_{c2}$, we calculate the angular velocity $\vect{\omega}_{i,j}$, which is induced by the virtual repulsive velocity $\vct{u}_{i,j}$ around an axis passing from $\vct{p}_i$ and defined by the cross product of the directions of $\vct{u}$ and  $\vct{p}_i$.
This angular velocity is given by:
\begin{equation}\label{eq:omega_pivot}
    \vect{\omega}_{i,j}\triangleq 
    \begin{cases}
          \frac{\vct{S}(\hat{\vct{p}}_{i,j}-\vct{p}_i)}{\|\hat{\vct{p}}_{i,j}-\vct{p}_i\|^2} \vct{u}_{i,j}, \; &\text{if} \; \frac{\vct{p}_i^\intercal}{\|\vct{p}_i\|} \vct{p}_{o,j} \in(0, \|\vct{p}_i\|) \\
          \vct{0}_3, \; &\text{otherwise}
    \end{cases}.
\end{equation} 


By summing the  $\vect{\omega}_{i,j}$-s acting on the $i$-th pivot point, we get the total angular veloity for the $i$-th OOI point, which is given by:
\begin{equation}\label{eq:omega_i}
    \vect{\omega}_{i} \triangleq \sum_{j=1}^m \vect{\omega}_{i,j} \in\mathbb{R}^3.
\end{equation}
Lastly, to synthesize the total control signal $\vct{v}_{c2}$, the $\vect{\omega}_i$-s are superimposed after calculating the corresponding linear velocity at the end-effector. This superimposition is given by:
\begin{equation}\label{eq:v_c2}
\vct{v}_{c2} 
     \triangleq k \sum_{i=1}^n  
     \begin{bmatrix}
        \vct{S}(\vct{p}_i) \\
        \vct{I}_3 
    \end{bmatrix}
     \vect{\omega}_{i} \in \mathbb{R}^6, 
\end{equation}
where $k\in\mathbb{R}_{>0}$ is a positive tunable gain.

Taking the time derivative of the artificial potential \eqref{eq:art_potential}, we find that:  $\dot{V}=-\frac{\|\vct{p}_i\|-\|\hat{\vct{p}}_{i,j}\|}{\|\vct{p}_i-\hat{\vct{p}}_{i,j}\|}\vct{u}_{i,j}^\intercal\vct{u}_{i,j}$, which is less or equal than 0, given that $\|\hat{\vct{p}}_{i,j}\|\leq\|\vct{p}_i\|$, which is true  by construction. This means that $V$ is bounded and  that $\hat{r}_{i,j}$ is increasing within the area of influence,  
due to the fact that $V(\hat{r}_{i,j})$ is a decreasing function of $\hat{r}_{i,j}$ (see Fig. \ref{fig:potential_vc2}). Therefore, $\vct{p}_i$ will not be occluded by the obstacle centered at $\vct{p}_{o,j}$. Furthermore, given that the linear velocity  of $\vct{v}_{c2}$ is orthogonal to $\vct{p}_i$, which implies that $\|\vct{p}_i\|$ remains constant, the maximum radius of visibility around $\vct{p}_i$ is increased with the proposed control signal. As a result,   the progressive unveiling of more points of the OOI  occurs in a chain-reaction manner, i.e. by unveiling progressively more and more OOI points.  




\section{Performance analysis through simulations}

To assess the performance of the proposed controller, which involves the superposition of the two body camera/end-effector velocities \eqref{eq:superimposition}, we tested it for 100 randomly 3D generated scenes with $m=5$ spherical obstacles that may occlude the OOI. We considered a linear segment as being the OOI. In particular, the procedure of the random scene generation is the following:

\begin{itemize}
    \item The five obstacle points are randomly placed within a cube with an edge of $0.5$m. In particular, this procedure can be described by:
     $   {}^0\vct{p}_{o,j}=
           \vct{U}_3^\intercal(-0.5,0.5) + [ 1.8 \; 
           -0.7 \;
           0.1]^\intercal
   $
    where $\vct{U}_3(-0.5,0.5)\in\mathbb{R}^3$ symbolizes the 3-dimensional random uniform process between $-0.5$ and $0.5$.
    \item The camera at the robot’s end-effector starts from the same initial pose, which is ${}^0\vct{p}_c = \vct{0}_3$ and
    $
        {}^0\vct{R}_c = [ 
           0 \; 0 \; 1; 
           1 \; 0 \; 0; 
           0 \; 1 \; 0 ]
    $.
    \item The linear segment which corresponds to the OOI is defined in a parametric form as:
    \begin{equation}
         {}^0\vct{p}_L(\sigma) = {}^0\vct{p}_0 + \sigma ( {}^0\vct{p}_1 - {}^0\vct{p}_0), 
    \end{equation}
    where ${}^0\vct{p}_0=[3 \; -0.1 \; 0.6]^\intercal$ and ${}^0\vct{p}_1=[3 \; 0.1 \; 1.4]^\intercal$ and $\sigma\in[0,\;1]$.
    
    \item The center of the ROI is $\vct{p}_r=0.5 (\vct{p}_0 + \vct{p}_1)$ and the radious is set to $r=3.262$, which means that the camera is initially within the ROI. 
\end{itemize}

\begin{figure}[!ht]
	\centering
	\begin{subfigure}[b]{0.23\textwidth}
		\centering
		\includegraphics[scale=0.605]{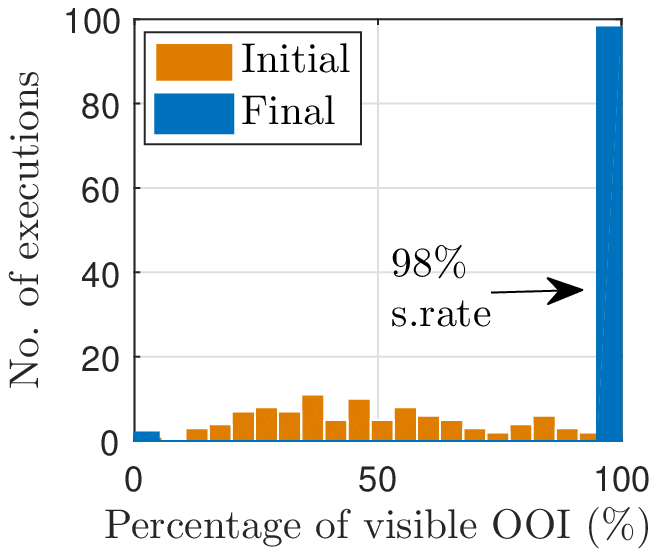}
		\subcaption{Success rate}\label{subfig:succes_rate}
	\end{subfigure}
	\begin{subfigure}[b]{0.23\textwidth}
		\centering
		\includegraphics[scale=0.6]{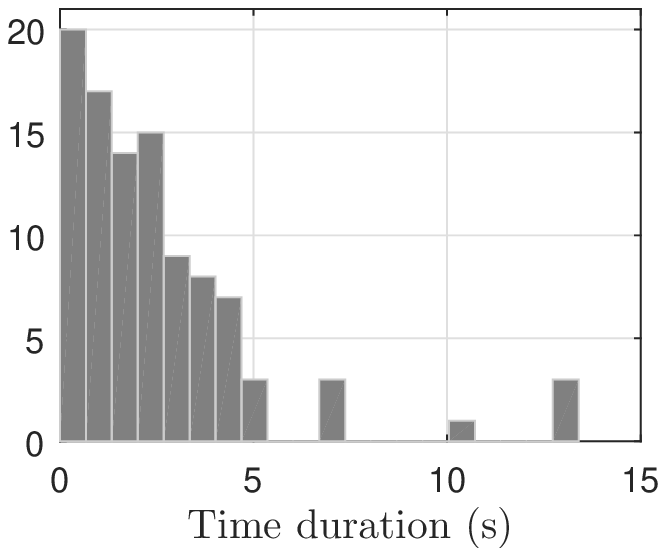}
		\subcaption{Duration}\label{subfig:duration}
	\end{subfigure}
	\caption{Histograms of the evaluation through 100 simulated scenes.}
	\label{fig:ismulation_results}
\end{figure}

The histogram of the initial visible percentage of visible OOI at the beginning of the simulations is depicted in Figure \ref{subfig:succes_rate}. Notice that the mean value of the initially visible portion of the OOI is 48.09\% for the 100 randomly generated scenes.


The parameters selected for the simulations are $d_c=0.1$, $d_0=0.3$, $k=0.04$, $k_p=5, k_o=200$, while the control signal is filtered via a first order low pass filter having a pole at $-10$.

The histogram of the visible percentage of the OOI at the end of the motion under the proposed control scheme is depicted in Figure \ref{subfig:succes_rate}. Considering as a success only the case where the whole OOI is visually unveiled, the success rate of the proposed controller reached 98\%. The histogram of the duration required for the motion is depicted in Figure \ref{subfig:duration} for all 100 simulation runs. It is clearly visible that the duration of motion is bounded, i.e. the system reaches an equilibrium and the mean duration of motion is 2.53s, with the selected values of the control parameters.

Through the simulation results it is evident that the control signal for the ROI reaching, i.e. $\vct{v}_{c1}$, does not reduce the efficiency of the controller which is responsible for the visual unveiling of the OOI, i.e.  $\vct{v}_{c2}$, while it is worth noting that the camera remained in the vicinity of the OOI and within the ROI during its motion, due to $\vct{v}_{c1}$.

\section{Experimental evaluation}

\begin{figure}[!ht]
	\centering
	\begin{subfigure}[t]{0.23\textwidth}
		\centering
		\includegraphics[scale=0.41]{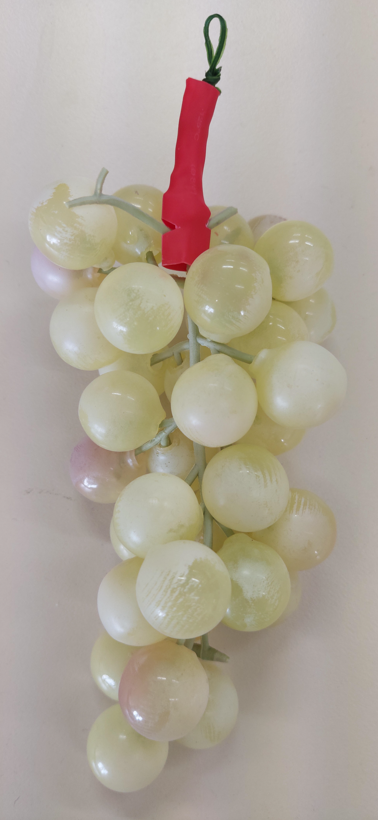}
		\subcaption{OOI: Grape stem, covered by a red tube for identification.}\label{subfig:grape}
	\end{subfigure}
	\begin{subfigure}[t]{0.23\textwidth}
		\centering
		\includegraphics[scale=0.39]{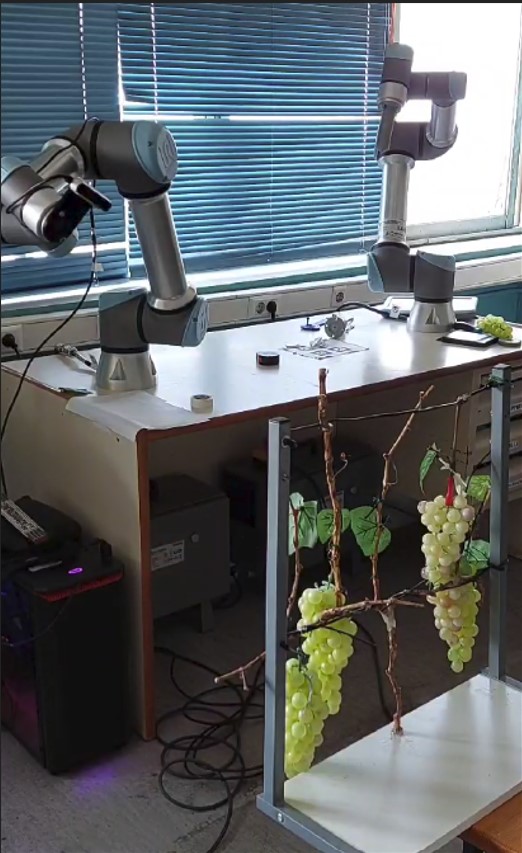}
		\subcaption{Mock-up vine setup.}\label{subfig:vine}
	\end{subfigure}
	\caption{Experimental setup.}
	\label{fig:exp_setup}
\end{figure}

For the experimental evaluation of the proposed method, a lab setup was built with a mock-up vine, involving real vine branches and plastic grape clusters, as illustrated in Figure \ref{fig:exp_setup}. We consider a pre-cutting scenario, in which the arm with the cutting tool is also equipped with an in-hand camera at its end-effector. Our aim is to approach the region of interest (ROI), which is defined as a sphere with radius $r=0.5$m around the grape cluster and visually unveil the stalk of the grape, which is considered to be the object of interest (OOI) for the task. In order to distinguish the points of the point-cloud which belong to the stalk (OOI), a red plastic tube was utilized (Figure \ref{subfig:grape}), as the identification of the stalk is out of the scope of this work.

An UR5e robot, fixed on a static desk, was utilized with a control cycle of 2ms. The RGB-D camera was capturing the RGB point-cloud with a framerate of 15fps, a resolution of 1280x720 pixels and the point-cloud was under-sampled 5 times for decreasing the total number of points. Furthermore, the control code for \eqref{eq:v_c2} was parallelized\footnote{The code parallelization involves the summation of \eqref{eq:v_c2}.}, employing 6 parallel threads of the CPU, utilizing the OpenMP library in C++. Owing to the parallel implementation, $\vct{v}_{c2}$ is generated with a control cycle of approximately 67ms. In contrast, the ROI reaching \& centering control signal, i.e. $\vct{v}_{c1}$, which does not generally involve a high computational load, had a control cycle of 2ms, i.e. it was synchronized with the control cycle of the robot. To superimpose these two control signals as dictated by \eqref{eq:superimposition}, sample-and-hold was utilized for $\vct{v}_{c2}$. The control parameters were selected to be $d_c=10^{-3}$m, $d_0=0.01$m, $k=0.05$ for \eqref{eq:u_ij}-\eqref{eq:v_c2}, $k_p=k_o=5$ for \eqref{eq:control_signal_rr} and a pole of $a=-2.5$ for the low pass filtering of the signal.

\begin{figure}[!ht]
	\centering
	\begin{subfigure}[b]{0.23\textwidth}
		\centering
		\includegraphics[scale=0.45]{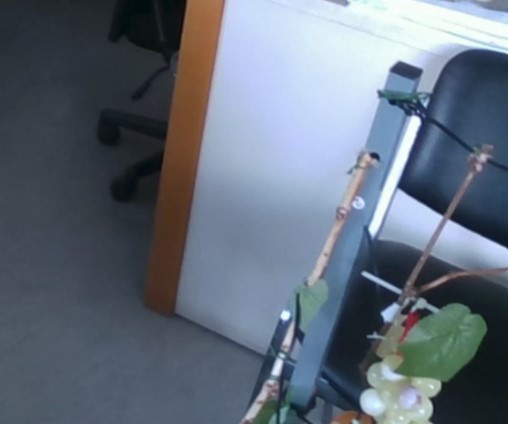}
		\subcaption{Initial.}\label{subfig:init_camera}
	\end{subfigure}
	\begin{subfigure}[b]{0.23\textwidth}
		\centering
		\includegraphics[scale=0.35]{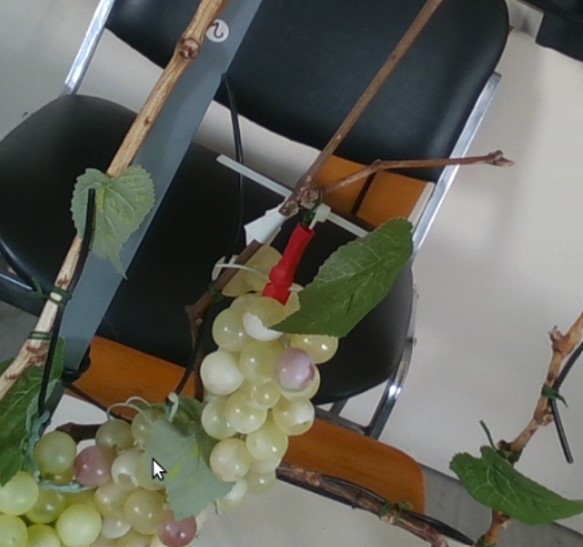}
		\subcaption{Final.}\label{subfig:final_camera}
	\end{subfigure}
	\caption{Initial and final camera view.}
	\label{fig:exp_camera_view}
\end{figure}

\begin{figure}[!ht]
	\centering
	\begin{subfigure}[t]{0.23\textwidth}
		\centering
		\includegraphics[scale=0.65]{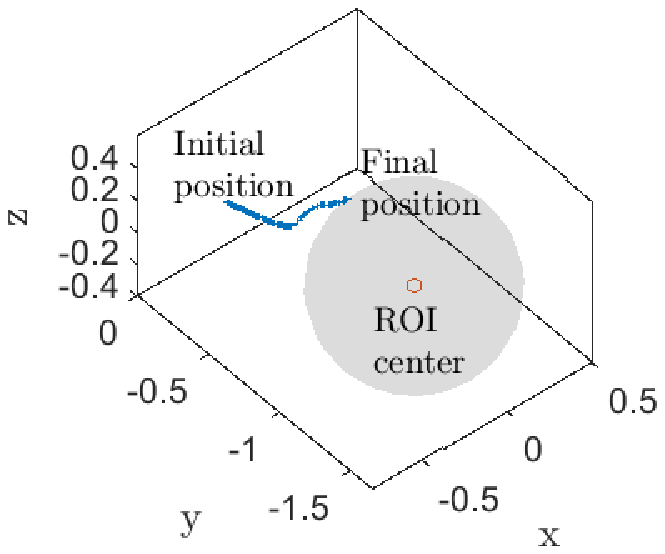}
		\subcaption{Path of the in-hand camera.}\label{fig:exp_path}
	\end{subfigure}
	\begin{subfigure}[t]{0.23\textwidth}
		\centering
		\includegraphics[scale=0.6]{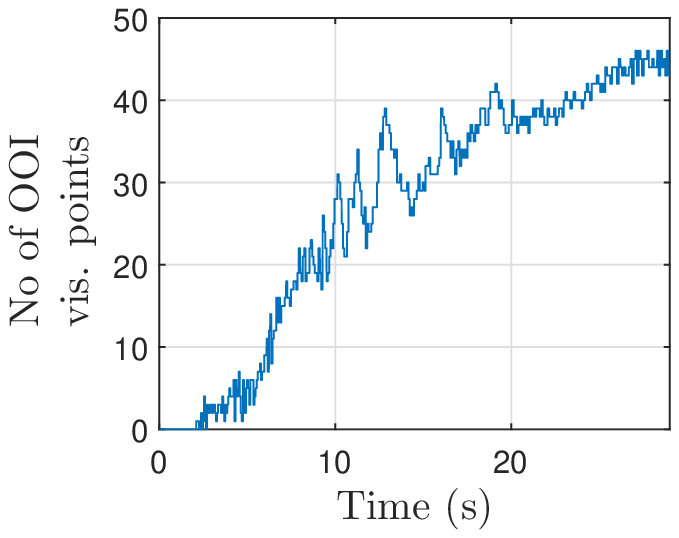}
		\subcaption{Number of visible points during the experiment.}\label{fig:vis_points}
	\end{subfigure}
	\caption{Experimental reults.}
	\label{fig:exp_results}
\end{figure}

The initial camera view of the stalk is shown in Figure \ref{subfig:init_camera}. Notice that, initially, only approximately 2-4 points of the stalk were visible to the system, due to the leaf occluding the view. The path followed by the camera at the end-effector with the proposed controller is shown in Figure \ref{fig:exp_path} and the final resulted camera view is provided in Figure \ref{subfig:final_camera}.

\begin{figure}[!htbp]
	\centering
	\includegraphics[scale=0.6]{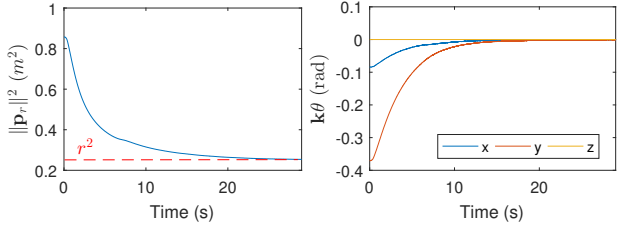}
	\caption{Distance from the center of ROI and centering orientation error. }
	\label{fig:rr_results}
\end{figure}

The number of visible points of OOI during the motion evolution, under the proposed control scheme, is depicted in Fig. \ref{fig:vis_points}. It is clearly visible that the number of visible points is increasing, which directly implies an increase at the visible area of the OOI from the camera, validating the results of the evaluations through simulations. In Figure \ref{fig:rr_results}, the time evolution of distance between the end-effector and the center of ROI is depicted, as well as the orientation error $\vct{k}\theta$ for centering of the view. Notice that the distance asymptotically reaches the radius of the ROI and the orientation error smoothly reaches zero, which demonstrates the fulfilment of the control objectives \ref{objective:region_reaching} and \ref{objective:centering} respectively.

Experimental results demonstrate that the control signal for the ROI reaching \& centering, i.e. $\vct{v}_{c1}$, and that responsible for the visual unveiling of the OOI, i.e. $\vct{v}_{c2}$, can be successfully superimposed achieving all set objectives  for ROI reaching \& centering and visual unveiling of the OOI.

\section{Conclusions}

This work addresses the problem of reaching and visually unveiling an object of interest by a  robot with a camera in hand based on the point cloud perceived by the camera. The models of the object of interest and the surrounding obstacles need not be known and there is no need of exploring the task space for unveiling the target.  Nevertheless, the proposed solution  assumes that the perception system  can on-line classify the points of the cloud to those belonging to the object of interest and the rest of objects. The proposed solution is synthesized by superimposing two body velocities for the camera/end-effector.
The control scheme effectiveness for reaching the vicinity of the object of interest while simultaneously optimizing the view of the object is revealed from both the simulation study and the experimental results. In particular, a success rate of 98\% is shown to be achieved in simulations  while the object of interest is shown to be significantly unveiled starting from just a few visible pixels, as demonstrated by the experimental results.   

\section*{APPENDIX: PROOF OF PROPOSITION 1}

Equations \eqref{eq:control_signal_rr}, \eqref{eq:ktheta} can be written in the world frame as follows:

\begin{equation} \label{eq:control_signal_rr_world}
    {}^0 \vct{v}_{c1}\triangleq 
    \begin{bmatrix}
       k_p\text{max}(0,\; f({}^0\vct{e}_p)){}^0\vct{e}_p \\
       k_o\theta {}^0\vct{k}
    \end{bmatrix},
\end{equation}
with ${}^0\vct{e}_p = {}^0\vct{p}_r - {}^0\vct{p}_c$, where ${}^0\vct{p}_c$ is the camera position and
\begin{equation}\label{eq:ktheta_world}
    \theta = \text{cos}^{-1}\left(\frac{\vct{z}^\intercal{}^0\vct{R}_c^\intercal{}^0\vct{e}_p}{\|{}^0\vct{e}_p\|}\right), \;
    {}^0\vct{k} = \frac{\vct{S}({}^0\vct{R}_c\vct{z}){}^0\vct{e}_p}{\|\vct{S}({}^0\vct{R}_c\vct{z}){}^0\vct{e}_p\|}
\end{equation}
with ${}^0\vct{R}_c$ the rotation matrix corresponding to the camera orientation. 

The kinematic system is  $[{}^0\dot{\vct{p}}_c^\intercal \; {}^0\vect{\omega}_c^\intercal]^\intercal = {}^0\vct{v}_{c1}$. We will now prove that the directions of ${}^0\vct{e}_p$ and ${}^0\vct{k}$ remain constant during the system evolution. Differentiating $\frac{{}^0\vct{e}_p}{\|{}^0\vct{e}_p\|}$ we get:
\begin{equation}\label{eq:deqhat}
    \frac{d}{dt}\frac{{}^0\vct{e}_p}{\|{}^0\vct{e}_p\|} = \frac{1}{\|{}^0\vct{e}_p\|}\left( \vct{I}_3 -  \frac{{}^0\vct{e}_p}{\|{}^0\vct{e}_p\|}\frac{{}^0\vct{e}_p^\intercal}{\|{}^0\vct{e}_p\|}\right){}^0\dot{\vct{e}_p}
\end{equation}
But since ${}^0\dot{\vct{e}_p}$ is in the direction of $\frac{{}^0\vct{e}_p}{\|{}^0\vct{e}_p\|}$ from (\ref{eq:control_signal_rr_world}), (\ref{eq:deqhat}) is equal to $\vct{0}$. Defining ${}^0\vct{a} \triangleq \vct{S}({}^0\vct{R}_c\vct{z}){}^0\vct{e}_p$, solving the second equation of (\ref{eq:ktheta_world}) for ${}^0\vct{a}$ and differentiating we get:
\begin{equation}\label{eq:dkappa}
    \|{}^0\vct{a}\|{}^0\dot{\vct{k}} + \frac{{}^0\dot{\vct{a}}^\intercal{}^0\vct{a}}{\|{}^0\vct{a}\|}{}^0\vct{k} = {}^0\dot{\vct{a}}
\end{equation}
But $\frac{{}^0\vct{a}}{\|{}^0\vct{a}\|} = {}^0\vct{k}$. We can also find that ${}^0\dot{\vct{a}}$ is also in the direction of ${}^0\vct{k}$. Therefore $\frac{{}^0\dot{\vct{a}}^\intercal{}^0\vct{a}}{\|{}^0\vct{a}\|}{}^0\vct{k} = {}^0\dot{\vct{a}}$ and thus from (\ref{eq:dkappa}) ${}^0\dot{\vct{k}}$ is also zero.
This implies that the second row of (\ref{eq:control_signal_rr_world}) which can be interpreted as the logarithmic orientation error, is the error between the camera orientation ${}^0\vct{R}_c$ and a constant orientation ${}^0\vct{R}_d$ such that $\theta{}^0\vct{k} = \log{}^0\vct{R}_d {}^0\vct{R}_c^\intercal$.
%

The angular velocity corresponding to the orientation error ${}^0\widetilde{\vct{R}} \triangleq {}^0\vct{R}_d {}^0\vct{R}_c^\intercal$ is equal to ${}^0\widetilde{\vect{\omega}} \triangleq {}^0\vect{\omega}_d - {}^0\widetilde{\vct{R}}{}^0\vect{\omega}_c$. Since ${}^0\vct{R}_d$ is constant, ${}^0\vect{\omega}_d = \vct{0}$. We can also utilize the Rodrigues' rotation formula to get ${}^0\widetilde{\vct{R}}{}^0\vect{\omega}_c = (\vct{I}_3 + \vct{S}({}^0\vct{k})(\sin\theta) + \vct{S}^2({}^0 \vct{k})(1-\cos\theta)){}^0\vect{\omega}_c= \vect{\omega}_c$. Therefore ${}^0\widetilde{\vect{\omega}}$ is equal to $-{}^0\vect{\omega}_c$. However the following equation holds \cite{leonidas_iros}:
\begin{equation}\label{eq:omegakth}
    {}^0\widetilde{\vect{\omega}} = \dot{\theta}{}^0\vct{k} + \sin{\theta}{}^0\dot{\vct{k}} + (1 - \cos{\theta})\vct{S}({}^0\vct{k}){}^0\dot{\vct{k}}
\end{equation}
Since ${}^0\dot{\vct{k}} = \vct{0}$, (\ref{eq:omegakth}) implies ${}^0\vect{\omega}_c = - \dot{\theta}{}^0\vct{k}$.
and thus:
\begin{equation}\label{eq:dth_th}
    \dot{\theta} = - k_o\theta
\end{equation}
Define the Lyapunov-like function $W$:
\begin{equation}\label{eq:Vlyapunov}
    W = \frac{1}{4} \max{\left(0, f({}^0\vct{e}_p)\right)^2} + \frac{1}{2}\theta^2
\end{equation}
Differentiating $W$ using (\ref{eq:control_signal_rr_world}) and (\ref{eq:dth_th}) yields:
\begin{equation}\label{eq:dVlyapunov}
    \dot{W} = -k_p \max{\left(0, f({}^0\vct{e}_p)\right)^2} {}^0\vct{e}_p^\intercal{}^0\vct{e}_p - k_o\theta^2
\end{equation}
Notice that $\dot{W}$ is zero, if and only if both the Objectives \ref{objective:region_reaching} and \ref{objective:centering} are satisfied and it is negative in all other cases.   
\bibliographystyle{IEEEtran}
\bibliography{mybib} 

\end{document}